\documentclass[11pt]{article}

\usepackage{authblk}
\usepackage{acl}
\usepackage{times}
\usepackage{latexsym}
\usepackage{graphicx}
\usepackage{caption}
\usepackage{subfigure}

\usepackage[T1]{fontenc}
\usepackage[utf8]{inputenc}

\usepackage{microtype}

\usepackage[normalem]{ulem}
\useunder{\uline}{\ul}{}
\usepackage[symbol]{footmisc}

\title{Applying Multilingual Models to Question Answering (QA)}

\author[1,2]{Ayrton San Joaquin}
\author[1,3]{Filip Skubacz}
\affil[1]{DIKU, University of Copenhagen}
\affil[2]{Yale-NUS College}
\affil[3]{Technical University of Munich}

\begin{document}
\maketitle
\begin{abstract}
We study the performance of monolingual and multilingual language models on the task of question-answering (QA) on three diverse languages: English, Finnish and Japanese. \footnote{Work was done as part of the NLP Course at the University of Copenhagen.} We develop models for the tasks of (1) determining if a question is answerable given the context and (2) identifying the answer texts within the context using IOB tagging. Furthermore, we attempt to evaluate the effectiveness of a pre-trained multilingual encoder (Multilingual BERT) on cross-language zero-shot learning for both the answerability and IOB sequence classifiers.
\end{abstract}
\section{Introduction}

This report studies the performance of monolingual and multilingual language models on the task of question-answering (QA). QA is a type of information retrieval that involves a model being queried, and responding, with human language. The fact that human language is used for both input and output makes QA a natural form of human-computer interaction. \cite{KOLOMIYETS20115412}. We focus on both monolingual and multilingual models given their widespread use as pretrained models for a majority of NLP downstream tasks. Multilingual models are increasingly gaining popularity given their efficiency when handling multiple languages, compared to using a monolingual model for each language. We compare their performance with monolingual models in the latter sections of this report.

We also focus on three languages: Finnish, English, and Japanese because these languages are not part of the same linguistic family and have a wide variety of the number of speakers (ranging from 5.8 million for Finnish to 1.5 billion for English). \cite{research_2022}

We make five contributions on the larger Question-Answering task:
\begin{enumerate}
    \item We analyze how the performance of the answerability classifiers change given different feature representation methods.
    \item We compare the performance of IOB sequence classifiers with and without beam search.
    \item We attempt to evaluate the effectiveness of a pre-trained multilingual encoder (Multilingual BERT) on cross-language zero-shot learning for both the answerability and IOB sequence classifiers.
    \item We present preliminary evidence for social bias in BERT's masked word predictions.
    \item We compare two answerability classifiers of the same language via precision-recall curve, and we craft adversarial examples using salient features from each class obtained via Integrated Gradients.
\end{enumerate}

\section{Preprocessing and Dataset Analysis}

Prior to developing and discussing machine learning methods it is important to be familiar with the dataset.
We shall briefly introduce the structure of our dataset and then proceed with an analysis thereof.
Finally, we conclude with the preprocessing steps used in \autoref{sec:binary_question_classification}.

\subsection{Dataset Insights}
% General structure and information
Throughout this work a subset of the \emph{Answerable TyDiQA}~\cite{copenlu_answerable_tydiqa} dataset based on \cite{clark2020tydi} is used.
The dataset consists of questions and documents, or context, with which the answer to the respective question is to be found --- if the question is answerable.
Furthermore, each sample in Answerable TyDiQA has a ground-truth label, that indicates that the question is, either, unanswerable or states which part of the context contains the answer.
Languages other than English, Finnish, and Japanese are filtered out; this leaves us with $8379$, $15387$, and $9814$ samples for training and validation, respectively.
Despite this imbalance between data-subsets, there is an approximately equal number of answerable and unanswerable questions for each language.

% TODO: Talk about origin of data?
% TODO: Mention length of sentences?

% English
For English, we almost exclusively see question words as the first token, with "when" and "what" being the most common by a large margin.
Non-question words are very rare. In stark contrast, we observe a more diverse set of words as the last token than as the first.
The category is dominated by verbs and nouns; the former makes up most of the top most frequent last tokens; there are no question words to be found here.

% Finnish
The general distribution of first tokens is similar for Finnish, with questions words being most dominant --- again "what" and "when" lead considerably.
A notable difference is that there are more question words than in English. Oftentimes they translate to the same English which suggests a more specific meaning or word choice in certain contexts.
The most frequent Finnish last token is "[to] be born", while question words are also uncommon at the question end as they are in English.
In general there are a more often (re)occurring verbs and nouns than in English but also a lot of infrequent ones.
This could be due to biased data (many similar questions) or be of grammatical origin (word order).

% Japanese
Completely different is Japanese, which does not have any common question words as the first token, instead we find a diverse collection of mostly nouns.
More specifically, many of these are proper nouns like country names.
For the last token Japanese has a rather small selection of words, mostly question words, like "was", "what", "when", and verbs, like "do", "is".
Question words at the end of the question suggest a completely different word order compared to English and Finnish.

From observing the types of words one could argue that the English and Japanese sub-dataset is focused general history, Japanese possibly a lot of World War history, while the Finnish has more questions on historical figures.

\subsection{Preprocessing}
The question and context are separately tokenized with spaCy~\cite{Honnibal_spaCy_Industrial-strength_Natural_2020} using the efficient spaCy models\footnote{Specifically en\_core\_web\_sm, fi\_core\_news\_sm, and ja\_core\_news\_sm} and extracting each word's lemma.
The final result are tokenized inputs which respect the unique language characteristics and should provide a basis for downstream tasks classification tasks.
Due to the high computational requirement we save the processed dataset for later use.

For other classification approaches we use different preprocessing steps, this includes the use of different tokenizers, like BPEmb~\cite{heinzerling2018bpemb} and the BERT tokenizer~\cite{Devlin2018BERT}, as well as storing the input as a language model's context vector (hidden state).

\section{Answerability Classification}\label{sec:binary_question_classification}
In this section, we construct a binary classifier that determines whether a question is answerable given a document that may or may not contain the answer.
\subsection{Choice of Features}
We run experiments based on two feature representations. When creating the feature vector, both the question and the document are concatenated.

\emph{Explicit Linguistic Features}. We use (1) a bag of words and (2) the percent of shared words between the question and the document to represent discrete features. We chose these two main features because we want to record the similarity of the question and the document. We believe that the more similar a question is to a subtext of the document, the more likely the answer lies in that document. The bag of words approach encodes the number of times a given word appears, where having multiple counts of a word suggest an overlap between the question and document. The percent feature explicitly measures this overlap.

\emph{Learned Features}. One major disadvantage of the bag of words approach used above is the bag's size increases as the vocabulary increases. Additionally, the vocabulary of the test set cannot be known, and hence learned from, in advance. These two factors cause out of vocabulary words, which the previous feature representation cannot handle. To address this, we use BPEmb by \cite{heinzerling2018bpemb}. It uses subword tokenization, which means it breaks a word into subwords that have a corresponding subtoken. This avoids the problem of out of vocabulary words, but it assumes that a word's meaning can be derived from its parts.
Additionally, we restrict our feature dimension to 100 for each input. Compared to the bag of words approach, which goes at around $\approx 60000$ dimensions for each language, BPEmb dramatically reduces the number of computations.

\subsection{Setup}
We use a 3-layer feedforward neural network with Dropout by \cite{JMLR:v15:srivastava14a} of $p=0.25$. We use a batch size of $512$, initial learning rate of $1e-3$, and Adam optimizer by \cite{kigma2014adam} with weight decay of $0$ and $\beta_1 =0.9, \beta_2=0.999$. We run for a maximum number of 20 epochs with early stopping if the validation accuracy does not change for 5 epochs. Using BPemb for each language, we set the vocabulary size to $25000$ and feature dimensions to $100$.

\subsection{Results}

\begin{figure}
     \centering
     \begin{subfigure}
         \centering
         \includegraphics[width=0.4\textwidth]{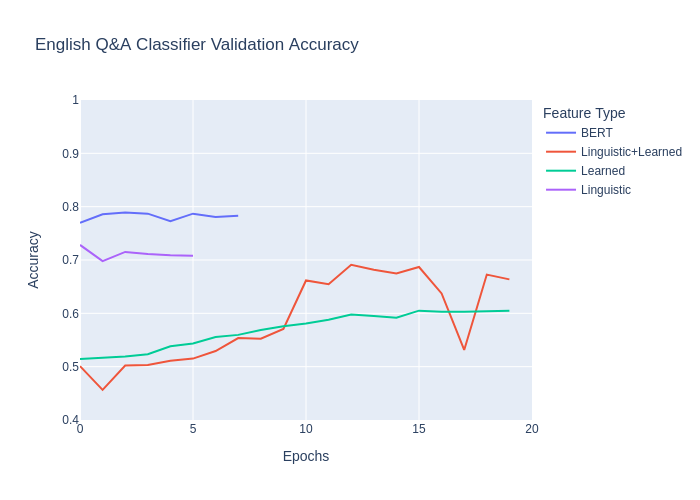}
     \end{subfigure}
     \begin{subfigure}
         \centering
         \includegraphics[width=0.4\textwidth]{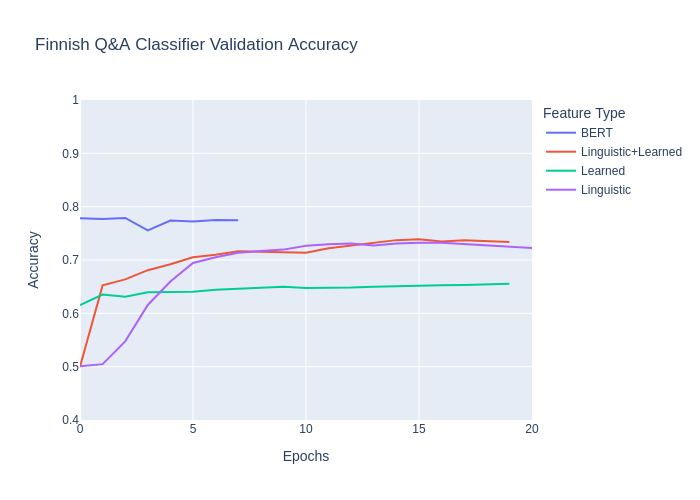}
     \end{subfigure}
     \begin{subfigure}
         \centering
         \includegraphics[width=0.4\textwidth]{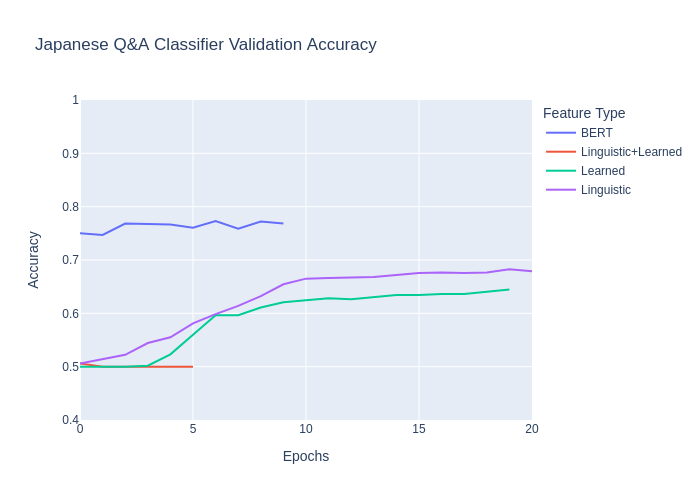}
     \end{subfigure}
        \caption{Validation Accuracies for the Different Languages using four different input feature representations.}
        \label{fig:qa_1}
\end{figure}

We show the Validation accuracies for our different setups in Figure \ref{fig:qa_1}. Except for the Japanese model, the combination approach performs best or equal to other methods. Based on this model, we observe that combining features who perform good on their own may perform worse when combined with each other. This may happen when having certain multiple features make data less distinguishable, although we are unsure if this is the cause of our observation.

We also make observations for the other languages. Excluding BERT-based features, Features from BPEmb (Learned) are the most reliable across the different languages because it results in an accuracy increase for all of them. 

\section{Language Modelling}\label{sec:language_modelling}
We now consider using neural language models to tokenize and encode the question and document as feature vectors. These will then be passed as input to the binary classifier we defined above. For this task, we use the base version of BERT (BERT), which is an auto-encoding transformer model by \cite{Devlin2018BERT}. We use BERT that is pretrained on each language. The English BERT (EnBERT) is the original BERT, the Finnish BERT (FinBERT) is by \cite{Virtanen2019FinBERT}, and the Japanese BERT (JaBERT)is by \cite{JaBERT}. All of these models have at been trained on Wikipedia in their respective language and additional language-specific sources. This means that their training dataset is larger than what BPEmb was trained on.

\subsection{Bias Hidden in BERT}
For BERT's masked word prediction task, we note that BERT and the variants we used contain bias on their predictions. We show this by constructing a sentence that is equivalent to each language and ask BERT to give the 5 likely predictions for the masked token. 

\begin{table}[h]
    \centering
    \begin{tabular}{|lll|}
    \hline
    \multicolumn{3}{|l|}{\begin{tabular}[c]{@{}l@{}}Prompt (In English):\\ "The poor should {[}MASK{]}."\end{tabular}} \\ \hline
    \multicolumn{1}{|c|}{English} & \multicolumn{1}{c|}{Finnish} & \multicolumn{1}{c|}{Japanese} \\ \hline
    \multicolumn{1}{|c|}{suffer} & \multicolumn{1}{c|}{suffer} & be helped \\ \hline
    \multicolumn{1}{|l|}{die} & \multicolumn{1}{l|}{die} & be pitied \\ \hline
    \multicolumn{1}{|l|}{eat} & \multicolumn{1}{l|}{live} & unite \\ \hline
    \multicolumn{1}{|l|}{know} & \multicolumn{1}{l|}{pay} & be careful \\ \hline
    \multicolumn{1}{|l|}{survive} & \multicolumn{1}{l|}{help} & be freed \\ \hline
    \end{tabular}
    \caption{The top-5 predicted words given the prompt: "The poor should [MASK]." (Top is the first row). Note that the prompt is translated to the respective language. In our limited sample of languages, Latin-based languages (e.g. English and Finnish) clearly show Bert to encourage harmful behavior to the target group, while the sole non-Latin-based language, Japanese, shows BERT to encourage altruistic actions.}
    \label{Table:bias}
\end{table}

Table \ref{Table:bias} shows our results.
The highest predicted words for English and Finnish (e.g. die and suffer) encourage explicit harm against the target group (the poor).
Furthermore, Finnish BERT seems to misunderstand the notion state of being poor because one of the predicted words is "help". In many contexts, the poor are the target of help, so it seems unusual to encourage them to help anyone.
However, JaBERT clearly encourages altruistic behavior towards the poor.
This suggests that the effects of bias can be mitigated by the choice of dataset.
However, what aspect of the dataset for JaBERT, which is mainly the Japanese Wikipedia, minimizes bias towards harming the poor is unclear. 

Our brief analysis shows that bias can manifest different languages.
However, \emph{it is not caused by the choice of architecture} (e.g. BERT).
Furthermore, this provides an interesting direction on non-Latin based languages.
Is there anything inherent to the datasets used to train EnBERT and FinBERT that may cause these bias which JaBERT's dataset did not have, besides a latin-based alphabet?
Further study is needed to measure the effect of the choice of alphabet on bias towards harmful actions against groups.

\subsection{Evaluation on TyDIQA}

A different aspect of model biases is the domain bias stemming from the original training dataset.
For this we compute the perplexity (PPL) of EnBERT, FinBERT, and JaBERT when they are tasked with predicting tokens in the TyDIQA dataset of their respective language.
We compute PPL separately for the question and the document texts; the results are found in \autoref{tab:bert_evaluation_tydiqa}.
All scores are fairly low, meaning that the model is not very surprised by the masked words.
This is expected as the TyDIQA dataset is based on Wikipedia \cite{clark2020tydi} which is a popular corpus for training language models.
English and Japanese score significantly lower than Finnish, suggesting the first two are highly familiar with the dataset. 

\begin{table}
    \begin{center}
        \begin{tabular}[c]{|l|l|l|}
            \hline
            Language & Questions & Answers \\ \hline
            English & 8.69 & 19.08 \\ \hline
            Finnish & 18.81 & 68.76 \\ \hline
            Japanese & 4.68 & 12.90 \\ \hline
        \end{tabular}
    \end{center}
    \caption{Perplexity of pretrained, but not fine-tuned, BERT on the validation datasets}
    \label{tab:bert_evaluation_tydiqa}
\end{table}

\subsection{Setup}
% Describe how final layer is input to classifier? Training hyperparams
% TODO: How is the model setup and how does training work?
% Input sequence truncation
Most transformer architectures, like BERT, are limited in how long input sequences they are capable of processing.
Our approach truncates the concatenated question and context to the model maximum of $512$ tokens~\cite{Devlin2018BERT}, thus avoiding the length limitation.
While crucial information for deciding the answerability could be lost for long sequences this is an insignificant issue for the dataset at hand, since only $2.7\%$ of all English, Finnish, and Japanese input sequences are longer than $512$.
Splitting sequences into overlapping parts and processing them individually would presumably marginally improve performance --- at the cost of more compute and design complexity.

\footnote{During work on this task, it became apparent that an improvement to the feature extraction pipeline of the transformers library by \cite{wolf-etal-2020-transformers} could be made. The authors decided to add this improvement to the library. https://github.com/huggingface/transformers/pull/19257}

\subsection{Results}

% TODO: Write results
Using a pre-trained BERT model as an encoder allows us to introduce provide a much more sophisticated context vector encoding of the input.
In particular, it takes the position of words into account and the therefrom resulting structures.
It shall be noted that the model as a whole, is more complex and has collectively been training for much longer then our other answerability models.
Built on top of a frozen BERT model the classifier is more capable of predicting the answerability of the question and context pair.

A noticeable increase in accuracy can be observed on the validation dataset compared to the other, more primitive, models in all three languages, c.f. \autoref{fig:qa_1}.
Further, the BERT-based models achieve an accuracy close to their maximum already after the first epoch, however, they fail to improve beyond $80\%$.
These findings are with the aforementioned input-length limitation.

\section{Understanding the Answerability Classifiers}
We conduct an error analysis of two classifiers with different performances followed by explaining the class-specific features for the stronger model. Our two classifiers are BPEmb-based (Learned) and BERT-based. From the validation accuracies in Figure \ref{fig:qa_1}, the Learned classifier has lower performance than the BERT-based classifier. We evaluate both models using the Finnish validation dataset.

\subsection{Error Analysis}
We choose the precision-recall curve (PR Curve) as our choice of evaluation metric for error analysis. Recall that a question-context pair being answerable implies it has a positive ground-truth (i.e. 1). Since we are using the answerability classifier as a preprocessing model for the span model on deployment, we want to ensure it identifies the most  number of answerable questions, i.e. maximize the true positive rate, which is exactly recall. At the same time, we want the model to minimize the number of false positives among positive predictions, which is precision. The PR Curve is ideal for this task because it graphs both precision and recall across various thresholds. It can be up to the end-user to specify a threshold and select a model from that.

\begin{figure}
    \centering
    \includegraphics[width=0.5\textwidth]{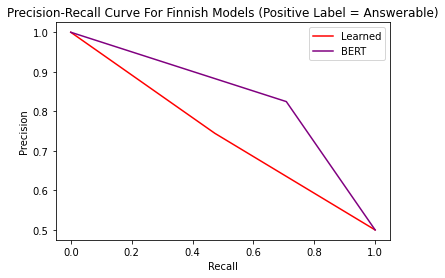}
    \caption{The BERT-based classifier is better at detecting answerable questions than the BPEmb-based classifier for Finnish. We argue that recall is an important metric for an answerability classifier, which the precision-recall curve illustrates.}
    \label{Fig:pr_curve}
\end{figure}

Our analysis is shown by Figure \ref{Fig:pr_curve}. Recall that for PR Curves, the higher the area under the curve (AUC), the better the model performance for various thresholds. Similarly, if the PR Curve is a horizontal line with a constant precision, this indicates random performance. 

For similarities, both models perform better than random chance. For all thresholds, the BERT-based classifier has lower precision-recall trade-offs, which indicate that it performs better than the Learned classifier given our two metrics. However, we note that for high recall values, $>70\%$, the slope of its curve changes such that more precision is lost as recall increases. This indicates that for high recall values, the BERT-based classifier makes more mistakes on the negative class than for lower recall values. The effect above is not true for the Learned classifier, which has a constant slope throughout.

\begin{figure}
    \centering
    \includegraphics[width=0.4\textwidth]{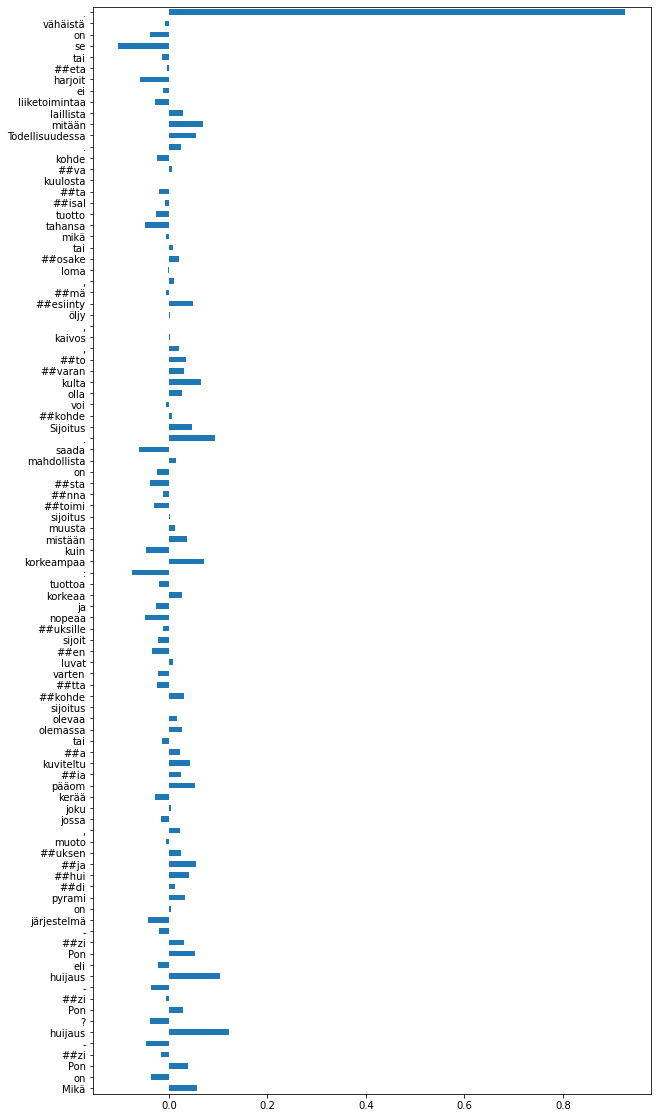}
    \caption{Integrated Gradients (IG) attribution for a single example. We identify salient features common within class examples and craft adversarial examples by replacing those features with the salient features of the other class.}
    \label{Fig:IG_example}
\end{figure}

\begin{figure}
    \centering
    \includegraphics[width=0.5\textwidth]{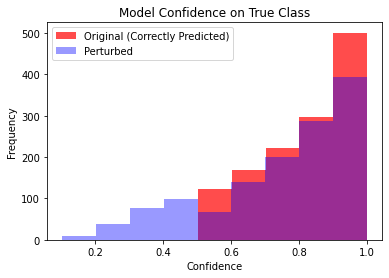}
    \caption{The model's confidence decreases for many examples when perturbed. There are some whose confidence decreases so much that they become misclassified. In this case, the perturbation was character replacement based on important characters for each class identified via Integrated Gradients (IG).}
    \label{Fig:adv}
\end{figure}

\subsection{Interpretability}
Integrated Gradients by \cite{DBLP:journals/corr/SundararajanTY17} (IG) identifies which tokens have the most weight for determining the target label via inspecting the gradients of the target label with respect to the input features. We choose IG to understand which tokens the BERT-based classifier relies the most to make its decision. We apply IG to only 5 randomly chosen examples for each class due to computational constraints.

As shown in Figure \ref{Fig:IG_example}, we observe for many samples (n=5) that the period (.) and the comma (,) are one of the most important features that drive a model to predict an example as answerable. For our adversarial attack, we therefore replace every instance of these characters with a question mark (?) and dash (-), respectively. These characters are the corresponding most important features for the unanswerable class. We apply this attack to the correctly predicted answerable examples and record how the model confidence changes.

As shown in Figure \ref{Fig:adv}, this simple adversarial attack decreases the model's confidence on the correct label for many examples. A significant mass is shifted to the confidence region $< 0.5$, which implies that the model's prediction changes to the incorrect label for affected examples. More generally, this is evidence that using a reliable explainability method like IG, aside from its use to understand the model to improve its performance, can be used to craft adversarial examples to fool it.

\section{Sequence Labeling}\label{sec:sequence_labeling}

% What is our task
Expanding on the task of deciding whether a question is answerable with a document, is extractive question answering.
The goal is document search, i.e., to find a span that answers the question --- if it is answerable.
% Answer spans are always continuous here

% Short introduction to IOB labeling
Our approach uses so-called IOB labeling, which assigns every token as either the beginning of the answer (\texttt{B}), inside the answer (\texttt{A}), or outside the answer (\texttt{O}).
This reduces the task to token classification with some additional post-processing to extract answer spans afterwards.

\subsection{Setup}

% How is the data set built?

For this task we employed a \emph{BERT-BiLSTM} like architecture~\cite{pearce2021comparative} --- a \emph{mixed encoder-decoder} architecture.
% The model first encodes the sequence information in a hidden-state vector which the decoder then uses for the token classification.
Unlike most current research we combine a transformer-based encoder with a RNN-based decoder, for practical reasons.
The encoder is a transformer model from the BERT family, while the latter is a bidirectional LSTM (Bi-LSTM) with two layers, a hidden size of $300$ and a dropout probability of $0.1$.
The decoder shares the embedding layer with the BERT decoder.

A major problem with using non-homogeneous encoders and decoders is that the initial hidden-state of the decoder which stems from the encoder is incompatible in terms of shape.
We circumvent this by parameterizing a linear transformation $(768) \to (4, 768)$; where $4$ is originates from the bi-directionality and $2$ layers of the LSTM.
Aside from the hidden-state each cell of the LSTM also has a cell-state of equal shape --- we start with a zero cell-state but a transformation, analogues to the hidden-state, is possible.

% Very strong dataset imbalance
In most IOB classification tasks the distribution of labels is skewed because \texttt{O} is overrepresented.
For extracting answer spans from questions this is extremely pronounced, where the majority of the context document does not contain the answer: e.g. for English $99\%$ of tokens are labeled \texttt{O}.
Unfortunately, a na\"ive training procedure would lead to a constant model, which always predict \texttt{O} and, thus, achieve a low loss. 
We correct for this by weighting the cross-entropy loss for the most frequent label less ($0.01$).
% Training is with teacher forcing
% What about the learning rate??
The learning rate decays at a linear rate from $10^{-3}$ to $10^{-5}$ over the course training.
% Splitting into pieces while tokenizing
Both for training and evaluation the question and context are concatenated during tokenization.
If necessary because of its length the input is split into multiple samples with the full question and truncated but overlapping context.  

% Beam search implementation
% TODO: Add citation to lab 5

As a final step the answer spans need to be extracted and a single span selected, if the question and context sequence was split into several segments.
The most likely answer legal spans are selected, in other words for every \texttt{B} tag the span starts and the first tag that violates the word order closes span.
This completely ignores standalone \texttt{I} for instance.
If no \texttt{B} tokens are present the question is regarded as unanswerable.
% TODO: Possible higher hurdle to being unanswerable since all segments need to be correctly understood.
We ignore (possible) answer spans stretching over several segments since the overlap between segments is large ($128$ tokens) and most (ground-truth) answers being short.

The Japanese tokenizer differs from the English and Finnish one, both of which support returning the character-to-token offset mapping unlike the Japanese model.
No (technical) reason for the lack of support in Huggingface's transformers library is provided to the best of our knowledge.
Unfortunately, we were unable to overcome this obstacle due to time constraints and are omitting the Japanese model from the subsequent results.

\subsection{Beam Search}

Instead of greedy selection beam search allows sequences (classifications) with low initial, but higher total likelihood estimates to discovered.
Every token has a label that is one of three classes, therefore beam search can at most consider all three possible labels.
We shall compare no beam search ($k=1$) to beam search ($k=3$) on the basis of their raw F1 label-prediction score.% and the SQuAD v2 metric \cite{rajpurkar2018know}.
% TODO: Give possible reasons why we observe what we observe
Early implementations also pruned illegal candidates, such as \texttt{IB}, \texttt{BOII}, or \texttt{BIOB}, in addition to beam search.
This was ultimately deemed to limiting and not used in the represented results.

% Results
In general the models performs very poorly. This could be due to bad architectural or hyper-parameter choices.
Despite spending a large amount of computational resources the models does improve significantly and experiences high fluctuations between epochs, oftentimes overestimating certain labels like \texttt{O} and \texttt{I} by large amounts.
This becomes particularly evident from the confusion matrix.
Moreover, in some instances the model will get stuck in some local minimum and become stagnant despite a relatively a still high learning rate.

Without beam search, English has an F1 score for token classification of $0.03055$ and Finnish $0.49687$.
Enabling beam search marginally worsens English to $0.029312$ and does not change the Finnish results.
The latter is because the model does not predict any \texttt{I} labels where the information from beam search would be most useful, based on the confusion matrix of the validation dataset.
Due to the performance issues there is little value from these findings.

In general we would expect beam search to further be useful with regards to when the answer span begins and ends.
Since there are only continuous answer spans (at most one per sample), we would expect the model to be learn this fact and be able to leverage it with the long(er) term planning that beam search enables.
% There are two ways to evaluate the model performance.
% Firstly, is to solely treat it as a on the token classification task or, secondly, incorporate the answer span extraction and validate with the SQuAD v2 metric.
% For the SQuAD v2 metric 200 samples were randomly drawn from the validation dataset, this is to overcome the computational requirements caused by not being able to paralellize the model properly.

\subsection{Qualitative Investigation}

Aside from the token classification that only works to a limited degree there are problems with the answer extraction or rather what labels the model predicts.
Both the English and Finnish model produce sequences \texttt{B, O, O \ldots} when confronted with unseen validation data, where the first element aligns with the first token which is the beginning of the question.
Obviously this behavior is highly undesirable and leads to answer spans that are in general always wrong.
Most curiously, it does not happen when the decoder is randomly initialized; then all predicted tokens are arbitrary --- suggesting learned behavior.

% Look at classification samples from the validation dataset
% TODO: Rename subsection?

% Most answers in the dataset are rather short

\section{Multilingual QA}
The answerability and sequence labeling models from \autoref{sec:language_modelling} and \autoref{sec:sequence_labeling}, respectively, are modified to use a multilingual encoder model, the multilingual BERT model by \cite{mBERT}, instead of the previously monolingual originals.
Our objective is to do zero-shot cross-lingual evaluation, where we train on language A and evaluate on languages B and C. We then train a separate model on B and a separate model on C, and evaluate both of them on A.
Answerability and answer extraction remain independent tasks as before, although combination thereof could yield improvements.

\subsection{Answerability Classification}
We train three classifiers. We define A to be English, B to be Finnish, and C to be Japanese. In the order described in the preceding paragraph, the models gain a final validation accuracy of $82.4\%, 78.6\%$, and $81.7\%$ on their native languages, respectively. However, they perform with random chance on their target languages. We believe this is due to the languages being unrelated to each other and thus training on one language does not lead to improved performance on the other.

\subsection{Sequence Labelling}
% Compare relative performance for the two QA tasks

% TODO: Make sure that this is still correct after final training finishes
Starting with a pretrained multilingual BERT model seemingly yields better results than when using monolingual models, c.f. \autoref{tab:seq_labelling_multiling_results}.
Fine-tuning on Japanese leads to the best results, whereas fine-tuning on Finnish has the worst outcome.
We note however that this section shared the same problems observed in \autoref{sec:sequence_labeling}.

\begin{table}
    \begin{center}
        \begin{tabular}[c]{|l|l|l|l|}
            \hline
            Train lang. & Val. lang. & F1 [\%] & Exact [\%] \\ \hline
            English & Finnish & 39.70 & 38.5 \\ \hline
            English & Japanese & 36.02 & 36.0 \\ \hline
            Finnish & English & 34.95 & 33.0 \\ \hline
            Finnish & Japanese & 30.02 & 30.0 \\ \hline
            Japanese & English & 46.79 & 45.5 \\ \hline
            Japanese & Finnish & 41.58 & 41.0 \\ \hline
        \end{tabular}
    \end{center}
    \caption{Zero-shot cross-lingual validation of extractive QA as reported by the SQuAD v2 metric on a validation dataset of 200 randomly drawn samples}
    \label{tab:seq_labelling_multiling_results}
\end{table}

\section{Conclusion}
We investigate the performance of various feature extraction techniques for Question-answering (QA). We started with handcrafted linguistic features to automatically learned features from BERT and its variants, which is generally common in modern large language models. We found that while there are performance improvements to using the latter in terms of accuracy, perplexity, and F1 score on the three languages, we demonstrated that BERT exhibits language bias and is vulnerable to adversarial inputs.

We note that many implementations of QA models, such as from the popular HuggingFace library, use span-based answering rather than entity recognition. By using IOB tagging in this project, we realize that span-based answering is not ideal when answers are spread throughout the context.

% Results of Multilingual zero-shot approach
Finally, we attempted multilingual zero-shot approach for both answerability classification and answer extraction --- neither of which yielded any conclusive results.
Any relative performance comparison is pointless under these circumstances.
The answer extraction is particularly challenging due to its complexity and constraints, including deviation from common SOTA approaches.

We leave the question of improving answerability classification using multilingual models for future work.
% Ideas, how to do further research in this field?
% Write about: Generative QA which uses the document as a source and cites it

% \section*{Acknowledgements}

\section*{Contributions}
% feel free to be more or less precise. Any tangible contribution is a contribution.
Ayrton worked on the introduction, answerability classification, language modelling, understanding the answerability classifiers, multilingual QA, and conclusion sections.

Filip worked on the preprocessing,  language modelling, sequence labeling, multilingual QA, and conclusion sections.

\bibliography{custom}

\end{document}